\documentclass[10pt,twocolumn,letterpaper]{article}

\usepackage{iccv}
\usepackage{times}
\usepackage{epsfig}
\usepackage{graphicx}
\usepackage{amsmath}
\usepackage{amssymb}
\usepackage[acronym]{glossaries}
\usepackage{booktabs}
\usepackage{adjustbox}
\usepackage{makecell}
\usepackage{multirow}
\usepackage{gensymb}
\usepackage{xcolor}
\usepackage{comment}
\usepackage{cite}
\usepackage[ruled,vlined]{algorithm2e}

\usepackage[pagebackref=true,breaklinks=true,letterpaper=true,colorlinks,bookmarks=false]{hyperref}

 \iccvfinalcopy 

\def\iccvPaperID{9} 

\newcommand*{\affaddr}[1]{#1} 
\newcommand*{\affmark}[1][*]{\textsuperscript{#1}}

\newcommand\blfootnote[1]{%
  \begingroup
  \renewcommand\thefootnote{}\footnote{#1}%
  \addtocounter{footnote}{-1}%
  \endgroup
}

\ificcvfinal\fi
\def\method{{MAPLE }}

\newcommand{\bs}{\boldsymbol}

\def\ie{{\it i.e.,\ }}
\def\etal{{et~al.}}

\def\vs{{\it vs.\ }}

\makeglossaries{}
\newacronym{tab}{Tab.}{Table}
\newacronym{fig}{Fig.}{Figure}
\newacronym{md}{MD}{Mahalanobis distance}
\newacronym{ood}{OOD}{Out-Of-Distribution}
\newacronym{sota}{SoA}{State-of-the-Art}
\newacronym{id}{ID}{In-Distribution}
\newacronym{ece}{ECE}{Expected Calibration Error}
\newacronym{nll}{NLL}{Negative Log-Likelihood}
\newacronym{auroc}{AUROC}{Area Under the Receiver Operating Characteristics}
\newacronym{aupr}{AUPR}{Area Under the Precision-Recall curve}
\newacronym{ed}{ED}{Euclidean Distance}
\newacronym{dnn}{DNN}{Deep Neural Network}
\newacronym{cnn}{CNN}{Convolutional Neural Network}

\begin{document}

\title{Gaussian Latent Representations for
Uncertainty Estimation using Mahalanobis
Distance in Deep Classifiers}

\author{Aishwarya Venkataramanan\affmark[1,2,3]\hspace{0.5em}
Assia Benbihi\affmark[4]\hspace{0.5em}
Martin Laviale\affmark[1,3]\hspace{0.5em}
C\'edric Pradalier\affmark[2,3]\hspace{0.5em}\\ \\
\affaddr{\affmark[1]Universit\'e de Lorraine, CNRS, LIEC, Metz, France }
\\ \affaddr{\affmark[2]Georgia Tech Europe, GT-CNRS IRL 2958, Metz, France} \\
\affaddr{\affmark[3]LTSER-``Zone Atelier Moselle", Metz, France} \\
\affaddr{\affmark[4]Czech Institute of Informatics, Robotics and Cybernetics, Czech Technical University in Prague}
}

\maketitle
\ificcvfinal\thispagestyle{empty}\fi

\blfootnote{Corresponding author: aishwarya.venkataramanan@univ-lorraine.fr} 

\begin{abstract}
Recent works show that the data distribution in a network's latent space is useful for estimating classification uncertainty and detecting \gls{ood} samples.
To obtain a well-regularized latent space that is conducive for uncertainty estimation, existing methods 
bring in significant changes to model architectures and training procedures.
In this paper, we present a lightweight and high-performance regularization method for \gls{md}-based uncertainty prediction, and that requires minimal changes to the network's architecture. To derive Gaussian latent representation favourable for \gls{md} calculation, we introduce a self-supervised representation learning method that separates in-class representations into multiple Gaussians. Classes with non-Gaussian representations are automatically identified and dynamically clustered into multiple new classes 
that are approximately Gaussian. Evaluation on standard \gls{ood} benchmarks shows that our method achieves state-of-the-art results on \gls{ood} detection and is very competitive on predictive probability calibration. Finally, we show the applicability of our method to a real-life computer vision use case on microorganism classification.
\end{abstract}

\section{Introduction}
\label{sec:intro}

Current deep learning classification networks achieve superior performance and find widespread applications in various industrial domains such as biology and robotics~\cite{litjens2017survey,grigorescu2020survey,senior2020improved}.
While they achieve state-of-the-art accuracy, there remain two main challenges that hinder the deployment of deep classifiers in critical situations: the derivation of calibrated classification and a measure of the classification uncertainty.
Without those, a network exposed to Out-of-Distribution (\gls{ood}) data makes incorrect predictions with high confidence~\cite{guo2017calibration} and no human-in-the-loop can catch such errors.
It is thus necessary to obtain calibrated probabilities~\cite{guo2017calibration} \ie predict probabilities that represent true likelihood, and to estimate the uncertainty in the network's predictions to allow users to make informed decisions.

\begin{figure}[t]
    \centering
    \begin{tabular}{cc}
    {Standard CNN}&{MAPLE}\\
    \includegraphics[width=0.45\linewidth,height=2.5cm]{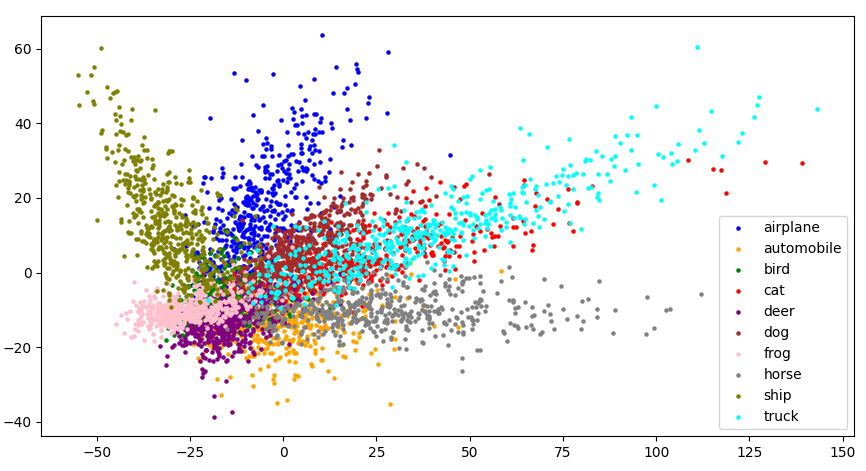} &
    \includegraphics[width=0.45\linewidth,height=2.5cm]{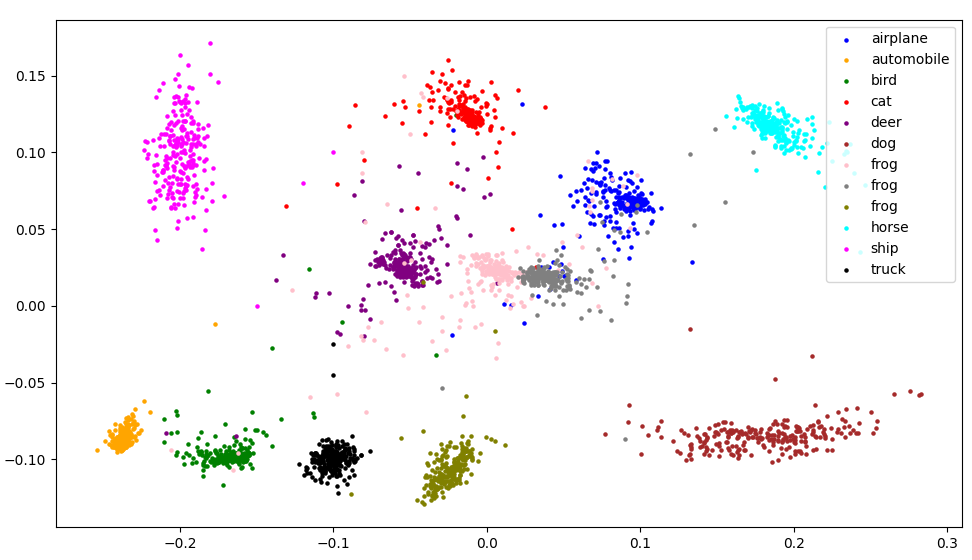}
     \\
    \end{tabular}
    \caption{\textbf{Self-supervised latent space regularization with \method} for uncertainty estimation and \gls{ood} detection. 
    \method improves class separation as illustrated by the PCA visualization of a CNN's latent space trained on CIFAR10 without regularization (left) and with \method regularization (right). Our method constrains the latent representations to be approximately Gaussian to enable efficient distance-based uncertainty estimation. 
    }
    \label{fig:latent}
\end{figure}

Among deep uncertainty estimation approaches~\cite{hullermeier2021aleatoric,gawlikowski2021survey,abdar2021review,gal2016uncertainty} are Bayesian Neural Networks~\cite{blundell2015weight}, MC-Dropout~\cite{gal2016dropout} and Deep Ensemble~\cite{lakshminarayanan2017simple}.
These stochastic methods require multiple forward-passes, so they are not scalable to large systems. Aware of the scalability requirements, current research focuses on estimating uncertainty from deterministic single-forward-pass networks~\cite{postels2020quantifying,liu2020energy,van2021feature,malinin2018predictive,hendrycks2016baseline,sensoy2018evidential}.
Distance-based methods belong to this category and are an attractive alternative for their excellent performance in \gls{ood} detection~\cite{van2020uncertainty,lee2018simple}.

Distance-based methods rely on the distance between the test samples and the \gls{id} samples in a network's latent space to determine if the test samples are \gls{ood}. A relevant distance is the Mahalanobis distance (MD)~\cite{mahalanobis} for its superior performance over \gls{ed}~\cite{vareldzhan2021anomaly,ren2021simple,kamoi2020mahalanobis}. One key \gls{md} assumption though is that the in-distribution samples in the latent space should follow class-conditional Gaussian distributions. In practice, though, there is nothing in the classification training that constrains the latent space to fulfil such an assumption~\cite{dinari2022variational}. Instead, research on representation learning shows that each class is usually composed of several 
clusters of visually similar
images~\cite{carbonnelle2020intraclass,venkataramanan2021tackling,em2017incorporating}. This can be due to intra-class variance of images taken from different view-points, the presence of additional objects in the image, and variations in object shapes. In the network's latent space, these variations appear as distinct distributions or deviate from a Gaussian distribution.
This breaks the \gls{md} assumption, which could lead to
incorrect or imprecise uncertainty estimation.
In this paper, we introduce MAPLE, a self-supervised representation learning method that regularizes a classification network's latent space to exhibit multivariate Gaussian distributions.
\method generates a latent space where class representations are Gaussian, making it compliant with the \gls{md} assumption and allows fast and high-performance \gls{md}-based \gls{ood} detection, uncertainty estimation, and calibrated classification.
The effect of \method is illustrated in Fig.~\ref{fig:latent} with the 2D projection of the latent space of a \gls{cnn} trained on CIFAR10.

\begin{figure}
    \centering
    \includegraphics[width=\linewidth]{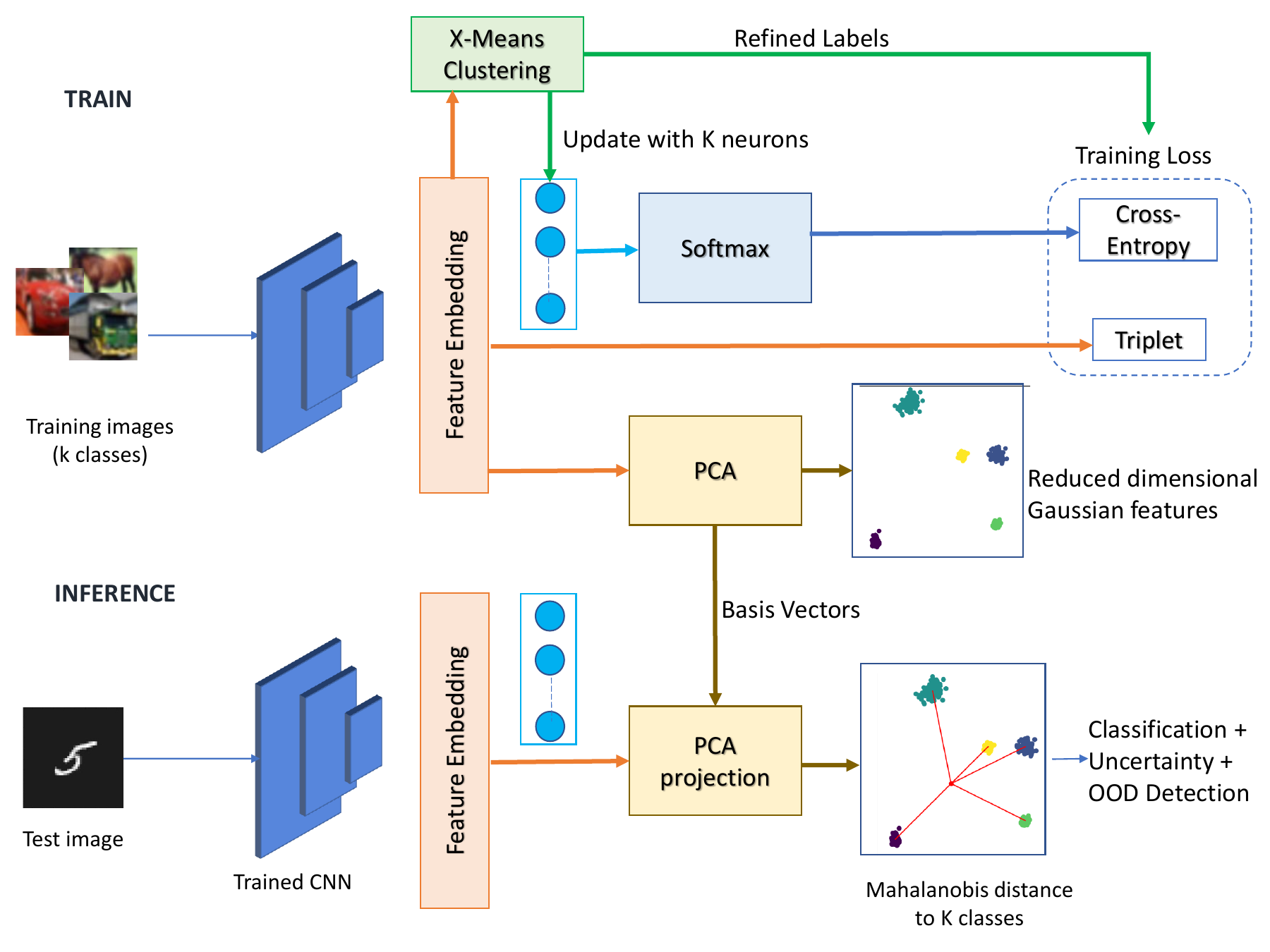}
    \caption{
    \textbf{Representation regularization with \method for uncertainty estimation.}
    Our approach trains a classification network to learn representations that are approximately Gaussian for each class.
    During inference, the Mahalanobis distance between a test sample and the class centroids is used for classification, uncertainty estimation and \gls{ood} detection.}
    \label{fig:pipeline}
\end{figure}

\method stands for MAhalanobis distance based uncertainty Prediction for reLiablE classification, and is illustrated in Fig.~\ref{fig:pipeline}. 
\method relies on two components: i) a self-supervised intra-class label refinement through clustering in the latent space; ii) a deep metric learning loss that improves the class separation.
During training, the representations associated to a class that deviate from a Gaussian distribution are divided into several clusters that are approximately Gaussian.
The cluster assignments become the new labels of the representations, and the training goes on. Since each cluster gathers samples that exhibit similar intra-class variations, the clustering step is akin to automatic fine-grained annotation. The metric-learning then reinforces the fined-grained class separation by pushing apart the new classes.
The combination of in-class clustering and metric learning results in classification representations that are well-clustered and approximately Gaussian, which makes them suitable for \gls{md}-based uncertainty estimation.

We evaluate \method against existing uncertainty quantification methods on the following standard benchmarks: CIFAR10~\cite{krizhevsky2009learning} \vs SVHN~\cite{netzer2011reading}/CIFAR100~\cite{krizhevsky2009learning}, CIFAR100 \vs CIFAR10/Tiny ImageNet~\cite{le2015tiny}, ImageNet~\cite{imagenet15russakovsky} \vs ImageNet-O~\cite{hendrycks2021natural} for \gls{ood} detection and predictive probability calibration. 
Results show that \method achieves the best compromise between performance and run time efficiency while being the most lightweight integration-wise.
Also, it introduces minor architectural changes and does not require additional fine-tuning to \gls{ood} datasets.

We summarize the paper's contributions as follows. \textbf{i)} 
We develop a self-supervised representation learning method that constrains a classification network's latent space to be approximately Gaussian.
\textbf{ii)} We show that such representations allow for reliable \gls{ood} detection and probability calibration using \gls{md}. 
\textbf{iii)} We design the method such that it has a minimal impact on the network's original architecture, and achieves results competitive with the state-of-the-art on \gls{ood} detection. 
The code is available in:~\url{https://github.com/vaishwarya96/MAPLE-uncertainty-estimation.git}
\section{Related Work} \label{sec:related_works}
\textbf{Multi-forward-pass Uncertainty Estimation.}
Traditional uncertainty quantification methods rely on Bayesian Neural Networks~\cite{goan2020bayesian,jospin2022hands} to learn a distribution over the network weights. To extract predictive probability variance, sampling~\cite{chen2014stochastic} or variational methods~\cite{blundell2015weight} are used.
The application of these methods is limited, as they increase the number of parameters by a factor of two and hinder convergence. 
As a lighter alternative, MC Dropout~\cite{gal2016dropout} enables dropout at test time and averages the network's output over several forward passes. While MC Dropout paves the way towards faster and lighter uncertainty estimation, it has been shown to produce over-confident predictions~\cite{lakshminarayanan2017simple} and underestimate uncertainty~\cite{smith2018understanding}. To improve uncertainty estimation, Deep Ensembles~\cite{lakshminarayanan2017simple} average the predictions from an ensemble of trained models and achieve state-of-the-art performance on several classification tasks. It remains computationally expensive due to the training of multiple models and the several forward passes during inference. 
By deriving uncertainty from a single forward pass, \method
achieves significantly faster inference time without sacrificing performance.

\textbf{Single-forward-pass Uncertainty Estimation.}
One line of work relies on the distribution of data samples in the network's latent space. A test sample is considered \gls{id} if it lies within the training data manifold, otherwise it is labelled as \gls{ood}. Methods differ in the way they regularize the representation space and the way they derive distances.
DUQ~\cite{van2020uncertainty} uses a Radial Basis Function (RBF) kernel in the representation space to measure distances between test samples and the centroids of various classes. Additionally, they use gradient penalty to obtain a regularized space, which improves the prediction's quality. SNGP~\cite{liu2020simple} uses Spectral Normalization on the network's weights to satisfy the bi-Lipchitz condition, which is a more gradient-friendly regularization than DUQ. This condition preserves semantically meaningful distance changes in the representation space with respect to input changes. The prediction's uncertainty is then given by a Gaussian Process layer on the output. To improve the scalability of the Gaussian Process estimation, \cite{van2021feature} proposes Deep Kernel Learning to process the input images with a distance-preserving network and fit a Gaussian on inducing points only.
Contrary to these methods, \method avoids the Gaussian Process estimation and gradient regularization during training and instead relies on simple metric learning.
Similarly, \cite{cheng2021learning} and VMDLS~\cite{dinari2022variational} simplifies the Gaussian enforcement by training the network with a Bregman divergence and KL-divergence loss respectively, so that each class representations follow an isotropic Gaussian distribution in the latent space for \gls{ood} detection. However, this ignores the possible intra-class variation within each class and requires the Gaussian variance to be tuned manually.
Instead, \method uses a simpler self-supervised clustering that automatically fits the data. Also, \method makes the latent space not only suitable for \gls{ood} detection but also for calibrated probability prediction.


\textbf{Mahalanobis-Distance for \gls{ood} detection.} \gls{md} is a common distance in the \gls{ood} detection literature. Early work by Lee \etal~\cite{lee2018simple} derives confidence values as a function of \gls{md} to predict the likelihood of a sample being \gls{id}.
To obtain competitive performance, 
the method requires several tweaks such as adding noise to input samples, combining confidence values from multiple feature layers, and fine-tuning on \gls{ood} datasets. \cite{kamoi2020mahalanobis} 
proposes two light improvements: Partial \gls{md} and Marginal \gls{md}. In Partial \gls{md}, the \gls{md} is computed on 
lower dimensional representations with PCA.
Marginal \gls{md} uses all training representations to fit a single Gaussian to calculate the \gls{md}. While both perform well on Far-\gls{ood} datasets \ie where \gls{id} and \gls{ood} samples are significantly distinct, their results are limited on Near-\gls{ood}~\cite{fort2021exploring}, where the \gls{ood} samples are semantically similar to the \gls{id} ones. Relative \gls{md} (RMD)~\cite{ren2021simple} improves the \gls{md} performance on Near-OOD by computing
a global MD between the test sample and the samples of all classes combined, and then subtracting this value from the per-class MDs. All these methods exhibit satisfying performance, but their main limitation
is their strong assumption that the image representations follow a Gaussian distribution, even though standard classification training does not enforce such a constraint. \method addresses this limitation with a self-supervised regularization.
By doing so, the features better fit the theoretical framework of \gls{md}-based \gls{ood} detection, thereby improving the performance.

\section{Method}

In this section, we describe MAPLE, a self-supervised regularization method for \gls{md}-based \gls{ood} detection, uncertainty estimation, and calibrated classification. 
It augments a standard \gls{cnn} classifier with a self-supervised regularization to output both class probabilities and \gls{md}-based uncertainty. To enable \gls{md} for \gls{ood} detection, the representations of the training samples are dynamically clustered into multiple Gaussians using X-Means~\cite{pelleg2000x} during training. The samples are assigned new pseudo-class labels defined by their cluster assignment. 
The network is then optimized with the cross-entropy loss and the triplet loss.
With periodic validation, the clusters are updated and the total number of classes change with every validation.
At inference time, the \gls{md} between a test sample and each cluster's centroid is used to estimate the classification uncertainty and the probability of the point being \gls{ood}.
Note that the only modification to the original network architecture is in the final layer, where the number of output neurons change according to the number of clusters identified. This makes \method easy to integrate to any classification network.
An algorithmic and computational description is provided in Appendix~\ref{app:algorithm}.

\subsection{Representation Regularization}
\label{sec:train}
This section describes the self-supervised automatic label refinement through clustering.

\textbf{Notations.}
Consider a classification problem consisting of $k$ classes with input samples $\boldsymbol{x}$ and labels $\boldsymbol{y}~\in~\{1,..,k\}$. The training dataset is denoted by
$\boldsymbol{\mathcal{D}_{train}}=\{(x_n,y_n)\}_{n=1}^{N}$ and the validation dataset by $\boldsymbol{\mathcal{D}_{val}}=\{(x_m,y_m)\}_{m=N+1}^{M}$. Let the training input samples be represented as $\boldsymbol{x_{train}}=\{x_n\}_{n=1}^{N}$. The training is done with an off-the-shelf classification \gls{cnn}. The penultimate layer of the CNN is used as a deep feature extractor $f^\theta(.)$, where $f^\theta:\mathbb{R}^D\rightarrow\mathbb{R}^d$ is a mapping from an input of dimension $D$ to a representation (or feature) of dimension $d$, and $\theta$ is the model's parameters.
The final layer consists of $k$ neurons, followed by softmax activation to obtain the predictive probabilities. In addition to the standard cross-entropy loss used in CNNs, we use the triplet loss on the representations to train the network. See Appendix.~\ref{app:loss} for a recall of these standard losses.

\begin{figure}
    \centering
    \includegraphics[width=\linewidth]{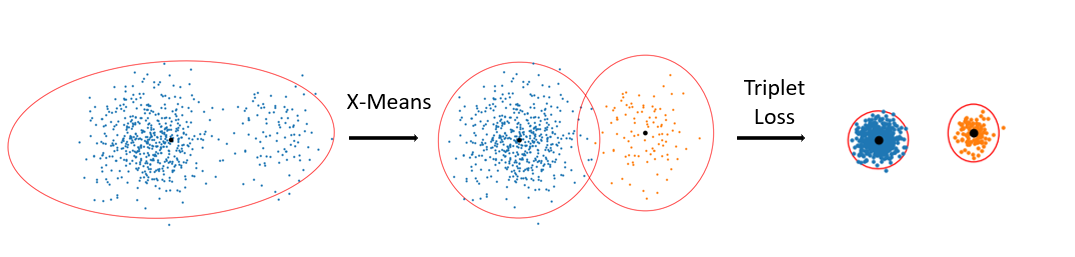}
    \caption{\textbf{Visualizing intra-class label refinement and feature optimization.}
    The original data is not perfectly Gaussian due to intra-class variations. X-Means refines the labelling by dividing the samples into multiple clusters that are approximately Gaussian. The clusters are considered as separate classes during training. 
    Triplet loss optimizes the representations by bringing the in-class samples together and separating them from other classes. 
    }
    \label{fig:cluster_example}
\end{figure}

\textbf{Self-Supervised Dynamic Relabelling.} During training, \method updates the training labels to make them representative of the features' separation in the latent space.
Every $p$ epochs, the network is evaluated on $\boldsymbol{\mathcal{D}_{val}}$ and the classes with a false negative ratio higher than a threshold $t$ are updated. This is representative of the scenarios where the samples of a given class are misclassified, which is typical of classes with high intra-class variations. For every class to update, the training representations belonging to such a class are extracted and clustered using X-Means~\cite{pelleg2000x}. The resulting clusters form well-separated groups and we use the cluster assigmnent as new pseudo-labels for the train samples. If $k^\prime$ additional clusters are introduced by X-Means, each of them are considered as independent classes. Thus, the number of classes becomes $K=k+k^\prime$, and the final layer of the model is updated to have $K$ neurons. Then, the network training continues with the new labels. 
During inference, the pseudo labels are remapped to the original set of $k$ labels to identify their original class.

Fig.~\ref{fig:cluster_example} illustrates the benefits of jointly using X-Means and the triplet loss on the representations: X-Means splits classes with high intra-class variations into separated classes that are semantically more representative of the data, and the triplet loss reinforces this separation.

The method introduces three hyperparameters: false negative ratio threshold $t$, frequency of validation epochs $p$ and the maximum number of clusters~(\texttt{max\_num\_cluster}), which is a parameter needed for X-Means.
More details on the hyperparameters are provided in Appendix.~\ref{app:hyp}.

\textbf{Clustering.}
The motivation for using X-Means over other commonly used clustering methods such as K-Means~\cite{lloyd1982least}, DB-SCAN~\cite{ester1996density} and Gaussian Mixture Models (GMMs) are two-folds: (1) X-Means is scalable and automatically identifies the number of clusters based on the Bayesian Information Criterion (BIC); (2) BIC uses a maximum likelihood estimation of the variance under the spherical Gaussian assumption, which means that the samples are approximately spherical Gaussian in each cluster.

\subsection{Representation Distance}
This section describes the \gls{md} derivation over the latent representations. To avoid matrix singularities, the latent representations are first reduced using PCA.

\textbf{Dimensionality reduction.} 
Representations extracted from large neural networks usually have a high dimension and redundant dimensions. The \gls{md} requires calculating the inverse covariance matrix of these features, but the presence of redundancy causes the covariance matrix to be singular. Furthermore, \cite{ren2021simple} shows that the presence of non-informative dimensions could be detrimental to \gls{md} performance. 
This motivates the use of dimensionality reduction. 

A common dimensionality reduction method is t-SNE~\cite{van2008visualizing}, widely used for latent space's visualization. While t-SNE maintains the local distribution of points, it fails to represent global distributions accurately, which is undesirable in distance-based uncertainty predictions. Instead, we use Principal Component Analysis (PCA) for dimensionality reduction.
The principal components are constructed from the covariance matrix of the standardized training representations. The eigen vectors of the covariance matrix are the principal components and the eigen values account for the amount of original information (variance) present in these components. 
We automatically estimate the number of principal components by the number of eigen values in decreasing order, required to explain 95\% of the original data variance. This transformation is denoted by $g:\mathbb{R}^d\rightarrow\mathbb{R}^{d^\prime}$, where $d^\prime$ is the dimension of the reduced features. 
With $\boldsymbol{x^\prime_{train}}=f^\theta(\boldsymbol{x_{train}})$ the full dimensions training features, we denote $\boldsymbol{z_{train}}=g(\boldsymbol{x^\prime_{train}})$ the reduced features. 

\textbf{Mahalanobis Distance.} The \gls{md} is a generalized version of Euclidean distance that takes into account the data correlation to measure the distance. Hence, the \gls{md} is more accurate when predicting the distance between a point and a distribution of points. Here, \gls{md} is calculated on the PCA-reduced representations as follows. Let $\{z_i\}$ be the set of training representations after dimensionality reduction, $\mu_c$ be the class centroids with $c=1,2,...,K$, and $\boldsymbol{\Sigma}$ be the shared covariance for all training samples, given by 
\begin{equation}
\begin{aligned}
\mu_c &= \frac{1}{N_c}\sum_{i:y_i=c} z_i \\ 
\boldsymbol{\Sigma} &= \frac{1}{N}\sum_c\sum_{i:y_i=K}(z_i-\mu_c)(z_i-\mu_c)^T
\end{aligned}
\end{equation}

The following Eq.~\ref{eq:md} gives the Mahalanobis distance between the centroid $\mu_c$ of class $c$ and a test sample ${\tilde x}$ with reduced representation ${\tilde z} = g(f^\theta({\tilde x}))$
\begin{equation}\label{eq:md}
    MD_c({\tilde x})= \sqrt{({\tilde{z}}-\mu_c)^T\boldsymbol{\Sigma}^{-1}({\tilde{z}}-\mu_c)}
\end{equation}

\subsection{Classification and Uncertainty Estimation}
\label{sec:uncertainty_estimation}
We now show how to use the \gls{md} distance calculated in Eq.~\ref{eq:md} for three purposes: classification, predictive probability, and uncertainty prediction. 

\textbf{\gls{md}-based Classification.} The predicted class is the one whose centroid $c^*$ is closest to the test sample $\tilde x$: 
\begin{equation}\label{eq:pred}
    c^* = \underset{c}{\text{argmin}}(MD_c({\tilde x}))
\end{equation}

Note that this classification is inferred in addition to the usual classification done by the network by taking the maximum of the output logits.

\textbf{Predictive Probability.}
We convert the \gls{md} into a calibrated classification probability using the following property: 
the squared \gls{md} on representations with dimension $d'$ follows a chi-squared distribution $\chi^2_{d^\prime}$ with $d^\prime$ degrees of freedom. The proof of this is provided in Appendix.~\ref{app:proof}.
The \gls{md} is converted as follows:
\begin{equation}
    P_{MD}^c = 1 - \text{cdf}(\chi^2_{d^\prime})(MD_c({\tilde x})^2)
\end{equation}
where cdf(.) is the cumulative distribution function. $P_{MD}^c$ represents the probability that a test sample belongs to class $c$. When the test point belongs to a particular class, the \gls{md} to that class is low and the corresponding $P_{MD}^c$ is high. The predictive probability is the one associated with the class $c^*$ obtained in Eq.~\ref{eq:pred}:
\begin{equation}
P_{MD}^{c^*} = \underset{c}{\text{max}}(P_{MD}^c)
\end{equation}

Note that contrary to a \gls{cnn} softmax `probabilities', this classification probability is calibrated and can be interpreted as a confidence in the classification output. This means $P_{MD}^c$ represents the actual probability that a sample belongs the class $c$.

\textbf{Uncertainty Prediction.} We define the predictive uncertainty, which is the uncertainty in the network prediction as
\begin{equation}
    u_{c^*} = 1 - P_{MD}^{c^*}
\end{equation}
For small values of \gls{md}, $u_{c^*}$ is around 0 and goes to 1 as the \gls{md} increases.

\section{Experiments}

We compare \method with the following related works: two multi forward-pass methods 
MC-Dropout~\cite{gal2016dropout} (10 dropout samples) and Deep ensemble~\cite{lakshminarayanan2017simple} (10 models), four single forward-pass methods: DUQ~\cite{van2020uncertainty}, SNGP~\cite{liu2020simple}, DUE~\cite{van2021feature} and VMDLS~\cite{dinari2022variational}. 
Following the standard evaluation on \gls{ood} detection, we evaluate the methods on classification, predictive probability calibration, and \gls{ood} detection on the following benchmark datasets: 
CIFAR10~\cite{krizhevsky2009learning} \vs SVHN~\cite{netzer2011reading}/CIFAR100~\cite{krizhevsky2009learning}, CIFAR100 \vs CIFAR10/Tiny ImageNet~\cite{le2015tiny} and ImageNet~\cite{imagenet15russakovsky} \vs ImageNet-O~\cite{hendrycks2021natural}. 

Additionally, we compare the \gls{id} metrics for the corrupted version of CIFAR100~\cite{hendrycks2021natural}. We also compare \method with \gls{md}-based methods on \gls{ood} detection, namely, the approach by Lee \etal~\cite{lee2018simple}, Marginal \gls{md}~\cite{kamoi2020mahalanobis} and R\gls{md}~\cite{ren2021simple}. We used the near-\gls{ood} CIFAR10 \vs CIFAR100 for the comparison, which is notably challenging for \gls{ood} detection.

\subsection{Evaluation Metrics}
We report the standard evaluation metrics~\cite{liu2020simple,van2020uncertainty}
namely, the classification accuracy, the \gls{ece}, the \gls{nll}, the \gls{auroc} and the \gls{aupr}. 
For qualitative analysis, we use calibration plots. 
As mentioned previously, \method produces two classification outputs, so we report the accuracies obtained from both the traditional softmax probability and the \gls{md}-based classification  (Sec.~\ref{sec:uncertainty_estimation}).
The \gls{ece} and the \gls{nll} are calculated from the predictive probability $P_{MD}^{c^*}$. \gls{auroc} and \gls{aupr} are calculated from the uncertainty $u_{c^*}$. 
The definition of these standard metrics are recalled in Appendix~\ref{app:metrics}.

\subsection{Implementation Details}
The CIFAR10 and CIFAR100 training follows
\cite{liu2020simple,van2021feature}
and uses a Wide ResNet 28-10~\cite{zagoruyko2016wide} for the classification backbone. 
The hyperparameters for the trainings are $p=10, t=0.3$ and \texttt{max\_num\_cluster}$=5$.
The ImageNet training is performed on ResNet-50~\cite{he2016deep}. The hyperparameters are $p=20, t=0.2$ and \texttt{max\_num\_cluster}$=5$.
Additional details on the dataset splits, hyperparameters, and the hardware used for training are provided in the Appendix~\ref{app:exp}.

\subsection{Results}
We report the results on CIFAR10, CIFAR100 and ImageNet in \gls{tab}~\ref{tab:cifar10_results}, \ref{tab:cifar100_results} and \ref{tab:imagenet_results} respectively.

\begin{table*}[h]
    \begin{adjustbox}{width=1.5\columnwidth, center}
    \centering
    \begin{tabular}{c||c c c|c c|c c|c}\hline
         \multirow{2}{*}{Method}&\multicolumn{3}{c|}{ID metrics}&\multicolumn{2}{c|}{OOD AUROC $\uparrow$}&\multicolumn{2}{c|}{OOD AUPR $\uparrow$}&{Latency$\downarrow$}\\
        &Accuracy $\uparrow$ & ECE $\downarrow$ & NLL $\downarrow$ & SVHN & CIFAR100 & SVHN & CIFAR100 &(ms/sample)  \\\hline
        Deterministic & 95.0$\pm$0.01 & 0.094 $\pm$0.002 & 0.138$\pm$0.01 & 0.801$\pm$0.01 & 0.765$\pm$0.01 & 0.794$\pm$0.01 & 0.762$\pm$0.01 & \textbf{4.01} \\
        MC Dropout~\cite{gal2016dropout} & 96.0$\pm$0.01 & 0.048$\pm$0.001 & 0.293$\pm$0.01 & 0.932$\pm$0.01 & 0.835$\pm$0.01 & 0.965$\pm$0.01 & 0.829$\pm$0.01&27.10\\  
        Deep Ensemble~\cite{lakshminarayanan2017simple} & \textbf{96.4$\pm$0.01} & 0.014$\pm$0.001 & \textbf{0.134$\pm$0.01} & 0.934$\pm$0.01 & 0.864$\pm$0.01 & 0.935$\pm$0.01 & 0.885$\pm$0.01 & 38.10\\ 
        DUQ~\cite{van2020uncertainty} & 94.5$\pm$0.02 & 0.023$\pm$0.001 & 0.222$\pm$0.01 & 0.927$\pm$0.01 & 0.872$\pm$0.01 & 0.973$\pm$0.01 & 0.833$\pm$0.01 &8.68\\ 
        SNGP~\cite{liu2020simple} & 95.7$\pm$0.01 & 0.016$\pm$0.001 & 0.153$\pm$0.01 & 0.991$\pm$0.01 & 0.911$\pm$0.01 & 0.994$\pm$0.01 & 0.907$\pm$0.01 &6.25\\ 
        DUE~\cite{van2021feature} & 95.6$\pm$0.02 & 0.015$\pm$0.001 & 0.179$\pm$0.01 & 0.936$\pm$0.01 & 0.852$\pm$0.01 & 0.967$\pm$0.01 & 0.850$\pm$0.01 & 6.94\\ 
        VMDLS~\cite{dinari2022variational} & 95.1$\pm$0.01 & - & - & 0.932$\pm$0.01 & 0.868$\pm$0.01 & 0.953$\pm$0.01 & 0.864$\pm$0.01 & 5.61\\ 
        \method & \textcolor{blue}{95.6$\pm$0.01}/\textcolor{orange}{95.4$\pm$0.01}& \textbf{0.012$\pm$0.001} & 0.142$\pm$0.01 & \textbf{0.996$\pm$0.01} & \textbf{0.926$\pm$0.01} & \textbf{0.997$\pm$0.01} & \textbf{0.918$\pm$0.01} & 4.96\\ \hline
        
    \end{tabular}
    \end{adjustbox}
    \medskip
    \caption{\textbf{CIFAR10 (\gls{id}) vs SVHN/CIFAR100 (\gls{ood}).} Results are averaged over 10 seeds.
    \method outperforms all single and multi pass methods on \gls{ood} detection, 
    and is significantly faster.
    Classification with \method is very competitive with the state-of-the-art and the predicted probabilities are better calibrated.
    \textcolor{blue}{Blue}: classification based on prediction from softmax probability.
    \textcolor{orange}{Orange}: \gls{md}-based classification.
    }
    \label{tab:cifar10_results}
\end{table*}

\begin{table*}[h]
    \begin{adjustbox}{width=1.8\columnwidth, center}
    \centering
    \begin{tabular}{c||c c| c c| c c| c c| c c}\hline
         \multirow{2}{*}{Method}&\multicolumn{2}{c|}{Accuracy$\uparrow$}&\multicolumn{2}{c|}{ECE$\downarrow$}&\multicolumn{2}{c|}{NLL$\downarrow$}&\multicolumn{2}{c|}{OOD AUROC$\uparrow$}&\multicolumn{2}{c}{OOD AUPR $\uparrow$}\\
        &Clean  & Corrupted  & Clean  & Corrupted & Clean & Corrupted & CIFAR10 & TinyImageNet & CIFAR10 & Tiny ImageNet \\\hline
        Deterministic & 79.0$\pm$0.02 & 52.2$\pm$0.03& 0.108$\pm$0.012 & 0.279$\pm$0.003& 1.342$\pm$0.03 & 2.834$\pm$0.03 & 0.697$\pm$0.01 & 0.748$\pm$0.01 & 0.713$\pm$0.01 & 0.747$\pm$0.01\\
        MC Dropout~\cite{gal2016dropout} & 79.4$\pm$0.02&46.3$\pm$0.05&0.115$\pm$0.010&0.293$\pm$0.004&0.986$\pm$0.02&2.868$\pm$0.02& 0.786$\pm$0.01 & 0.787$\pm$0.01&0.781$\pm$0.01 & 0.790$\pm$0.01\\
        Deep ensemble~\cite{lakshminarayanan2017simple} & \textbf{79.6$\pm$0.01}&{54.0$\pm$0.06}&\textbf{0.029$\pm$0.008}&0.254$\pm$0.005&\textbf{0.706$\pm$0.01}&2.893$\pm$0.02&\textbf{0.798$\pm$0.01}&0.811$\pm$0.01&0.792$\pm$0.01&0.801$\pm$0.01\\
        DUQ~\cite{van2020uncertainty} & 77.6$\pm$0.02 &50.5$\pm$0.04 &0.112$\pm$0.015&0.277$\pm$0.006&1.303$\pm$0.03&2.811$\pm$0.02&0.740$\pm$0.01&0.759$\pm$0.01&0.747$\pm$0.01&0.761$\pm$0.01\\
        SNGP~\cite{liu2020simple} & 78.7$\pm$0.01&50.5$\pm$0.03&0.129$\pm$0.012&0.286$\pm$0.003&1.080$\pm$0.01&\textbf{2.676$\pm$0.02}&0.743$\pm$0.01&0.783$\pm$0.01&0.749$\pm$0.01&0.765$\pm$0.01\\
        DUE~\cite{van2021feature} & 77.8$\pm$0.02&49.3$\pm$0.05&0.134$\pm$0.014&0.305$\pm$0.005&1.454$\pm$0.02&2.756$\pm$0.03&0.732$\pm$0.01&0.754$\pm$0.01&0.734$\pm$0.01&0.768$\pm$0.01\\
        \method &  \textcolor{blue}{78.9$\pm$0.02}/\textcolor{orange}{78.6$\pm$0.01}&\textbf{\textcolor{blue}{54.2$\pm$0.04}}/{\textcolor{orange}{54.0$\pm$0.03}}&0.065$\pm$0.001&\textbf{0.245$\pm$0.004}&1.112$\pm$0.01&2.715$\pm$0.02&0.793$\pm$0.01&\textbf{0.828$\pm$0.01}&\textbf{0.799$\pm$0.01}&\textbf{0.817$\pm$0.01} \\ \hline
    \end{tabular}
    \end{adjustbox}
    \medskip
    \caption{\textbf{CIFAR100 (ID) vs CIFAR10/Tiny ImageNet (\gls{ood}).} Results are averaged over 10 seeds. We also evaluate the ID metrics for the corrupted version of CIFAR100 from \cite{hendrycks2021natural}.
    \method achieves the best performance on \gls{ood} detection. It is very competitive with other single-pass methods on the classification task.
    \textcolor{blue}{Blue}: Classification based on prediction from softmax probability \textcolor{orange}{Orange}: \gls{md}-based classification.
    }
    \label{tab:cifar100_results}
\end{table*}

\begin{table}[h]
    \begin{adjustbox}{width=\columnwidth, center}
    \centering
    \begin{tabular}{c||c c c|c c | c}\hline
         \multirow{2}{*}{Method}&\multicolumn{3}{c|}{ID metrics}&\multicolumn{2}{c|}{OOD metrics}&{Latency $\downarrow$}\\
        &Accuracy $\uparrow$ & ECE $\downarrow$ & NLL $\downarrow$ & AUROC $\uparrow$& AUPR $\uparrow$ & {(ms/sample)}\\\hline
        Deterministic & 75.7$\pm$0.03 & 0.058$\pm$0.004 & 0.925$\pm$0.03 & 0.553$\pm$0.01 & 0.546$\pm$0.01 & \textbf{42.64} \\
        MC Dropout~\cite{gal2016dropout} & 75.3$\pm$0.04 & 0.032$\pm$0.003 & \textbf{0.922$\pm$0.02} & 0.614$\pm$0.08 & 0.609$\pm$0.05 & 93.74\\  
        Deep ensemble (3 models)~\cite{lakshminarayanan2017simple} & \textbf{76.4$\pm$0.08} & 0.024$\pm$0.004 & 0.930$\pm$0.03 & 0.625$\pm$0.05 & 0.616$\pm$0.04 & 130.78 \\ 
        \method &  \textcolor{blue}{75.6$\pm$0.05}/\textcolor{orange}{75.2$\pm$0.07}&\textbf{0.021$\pm$0.003}&0.928$\pm$0.03&\textbf{0.637$\pm$0.06}&\textbf{0.635$\pm$0.04} & 55.26\\ \hline
    \end{tabular}
    \end{adjustbox}
    \medskip
    \caption{\textbf{ImageNet (ID) vs ImageNet-O (\gls{ood}).} Results are averaged over 10 seeds. MC-Dropout is performed for 10 forward passes.
    \method achieves the best performance on \gls{ood} detection. It is very competitive with other single-pass methods on the classification task.
    \textcolor{blue}{Blue}: Classification based on prediction from softmax probability \textcolor{orange}{Orange}: \gls{md}-based classification.
    }
    \label{tab:imagenet_results}
\end{table}

\textbf{OOD Detection Results.} \method outperforms the baseline methods by upto 12\% on \gls{ood} detection. 
Note that competitive approaches, such as SNGP and DUE, derive their performance from spectral normalization and Gaussian process layer, which are invasive training add-ons. In contrast, \method relies only on the layers of
a standard \gls{cnn} architecture to achieve superior performance. 


\textbf{Classification Results.} \method achieves results competitive to state-of-the-art, only 1\% below the top method Deep ensemble~\cite{lakshminarayanan2017simple} whose score comes at the cost of training and inference on several models.
Note that both \method accuracies, the softmax probability and the \gls{md}-based one are close. A finer analysis of the accuracy shows that the slight difference in accuracy with the \gls{md}-based classification occurs on samples the network is uncertain about: \method achieves top accuracy on high-confidence predictions (above 80\% and 90\% confidence) and the accuracy slightly decreases for lower-confidence predictions. See Appendix.~\ref{app:accuracy} for an extended analysis.

\begin{figure}[h]
    \centering
    \begin{tabular}{cc}
    \includegraphics[width=0.45\linewidth]{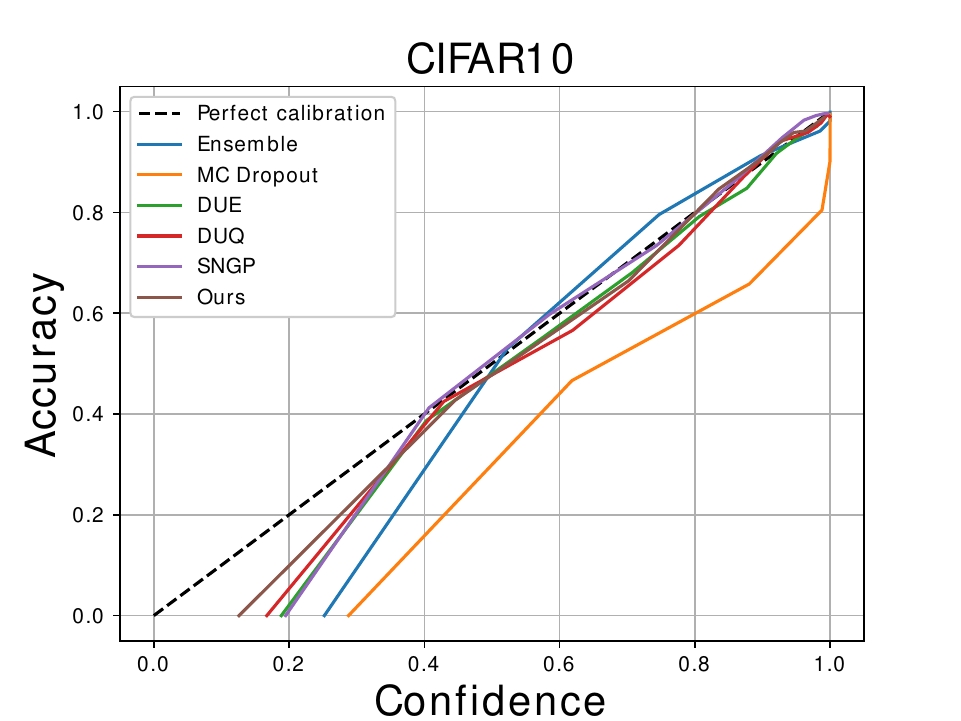} &
    \includegraphics[width=0.45\linewidth]{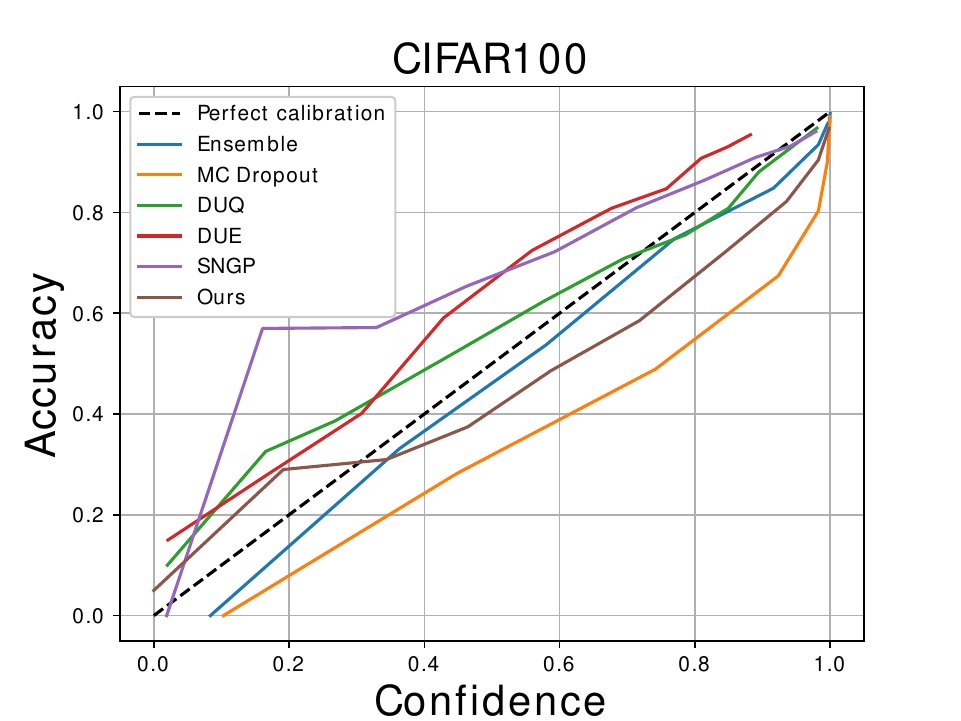} 
     \\
    \end{tabular}
    \caption{\textbf{Calibration plots.} A perfectly calibrated plot is when the predicted confidence equals the true likelihood \ie the accuracy. This is shown by the linear dotted line in the plots. } 
    \label{fig:calibration}
\end{figure}

\textbf{Calibration Results.} \method is competitive with state-of-the-art SNGP~\cite{liu2020simple} and Deep Ensembles. 
The calibration plot for CIFAR10 and CIFAR100 and shown in Fig.~\ref{fig:calibration}.
On CIFAR10, all methods are well-calibrated, except for 
MC-Dropout that is overconfident in its predictions, 
which explains its high ECE score. When the accuracy is below 0.4, baseline methods become overconfident,
whereas \method is closer to optimal calibration and achieves the best \gls{ece} score.

Additional results on Gaussian test, are provided in Appendix.~\ref{app:additional_exp}.

\subsection{Comparison with other \gls{md} methods}

\textbf{Setup.} \method is compared against \gls{md}-based \gls{ood} detectors~\cite{lee2018simple, kamoi2020mahalanobis,ren2021simple}. These methods are tailored for \gls{ood} detection, so we report the metric relevant to this task only for the sake of fairness. We report the \gls{auroc} score on the challenging near-\gls{ood} dataset CIFAR10 \vs CIFAR100. The experiments are done with a Wide ResNet 28-10~\cite{zagoruyko2016wide}.

\begin{table}[h]

    \begin{adjustbox}{width=\columnwidth, center}
    \centering
    \begin{tabular}{c c c c c }\hline
        {Method}&{Lee \etal~\cite{lee2018simple}}&Marginal \gls{md}~\cite{kamoi2020mahalanobis}&RMD~\cite{ren2021simple} & {Ours}\\\hline
        {AUROC $\uparrow$} & 0.893 & 0.838 & 0.897 & \textbf{0.926}  \\\hline
    \end{tabular}
    \end{adjustbox}
    \medskip
    \caption{\textbf{Comparison with \gls{md}-based \gls{ood} detection.}
    \method performs better in \gls{ood} detection than existing \gls{md}-based methods on the CIFAR10 \vs CIFAR100 setup. 
    }
    \label{tab:md_comparison}
\end{table}

\begin{table*}[h]

    \begin{adjustbox}{width=1.8\columnwidth, center}
    \centering
    \begin{tabular}{c||c c c|c c | c c| c}\hline
         \multirow{2}{*}{Method}&\multicolumn{3}{c|}{ID metrics}&\multicolumn{2}{c|}{OOD metrics - SVHN}&\multicolumn{2}{c|}{OOD metrics—CIFAR100}&\multirow{2}{*}{\#Eig}\\
        &Softmax Accuracy $\uparrow$&\gls{md}-based Accuracy$\uparrow$ & ECE $\downarrow$ & AUROC $\uparrow$ & AUPR $\uparrow$& AUROC $\uparrow$ & AUPR $\uparrow$\\\hline
        \textbf{DNN+MD (1)} &0.950&0.943&0.086&0.752&0.762&0.583&0.564&-\\
        \textbf{DNN+PCA+MD (2)} &0.950&0.946&0.053&0.855&0.839&0.813&0.859&12\\\hline
        \textbf{DNN+PCA+ED (3)} &0.950&0.943&0.105&0.829&0.804&0.734&0.765&12 \\ \hline
        \textbf{DNN+Triplet+PCA+MD (4)} &0.954&0.953&0.013&0.945&0.948&0.912&0.894&11\\
        \textbf{DNN+Clustering+PCA+MD (5)} &0.947&0.945&0.032&0.922&0.908&0.811&0.815&12\\\hline
        \textbf{\method (6)} & \textbf{0.956}&\textbf{0.954}&\textbf{0.012}&\textbf{0.996}&\textbf{0.997}&\textbf{0.926}&\textbf{0.930}&12\\ \hline
    \end{tabular}
    \end{adjustbox}
    \medskip
    \caption{\textbf{Ablation study.}
    We evaluate the influence of several \method components. 
    \textbf{PCA} (1 vs 2) results in a significant improvement of the \gls{ood} detection by discarding non-informative dimensions. The distances derived on these reduced features are better representative of the similarity between the input samples.
    The \textbf{\gls{md}} (2 vs 3) is better suited than \gls{ed} for calibrated classification and \gls{ood} detection, which reiterates conclusions already found in previous works. The \textbf{triplet loss} (2 vs 4) improves both the accuracy and the \gls{ood} metrics by increasing the class separation. \textbf{Clustering} alone (2 vs 5) also contributes to a better separation of the classes, but the results are not as significant. The joint use of \textbf{triplet loss and clustering}, as done in \method (6) achieves the best results on both classification and \gls{ood} detection. Note: \#Eig refers to the number of principal components, whenever applicable.
    }
    \label{tab:ablation}
\end{table*}

\textbf{\gls{ood} Detection Results.}
\method achieves top-performance on Near-\gls{ood} detection (\gls{tab}~\ref{tab:md_comparison}),
which supports MAPLE's representation regularization. Note that the primary difference between \method and the baselines is their lack of constraints on the latent representation. In contrast, we force the samples of every class to be Gaussian before calculating \gls{md}.
Non-Gaussian samples lead to incorrect mean and covariance calculations, resulting in incorrect distance values. The error is more pronounced when the samples deviate from the Gaussian distribution by a large factor. This explains why the \gls{md}-based approaches under-perform compared to \method on Near-OOD.

\subsection{Ablation analysis}
In this study, we assess how the different components of \method impact its performance.
We train a Wide ResNet 28-10~\cite{zagoruyko2016wide} network on CIFAR10 and use SVHN and CIFAR100 as \gls{ood} datasets. 

\textbf{Dimensionality Reduction.}
We consider two scenarios: 
\textbf{(1) DNN+MD} - A baseline where a standard \gls{dnn} is trained
with the cross-entropy loss and with no feature regularization.
The \gls{md} is computed on the raw features,
and we add a value of $1e^{-20}$ to the diagonal elements~\cite{ren2021simple} to avoid a singular covariance matrix.
\textbf{(2) DNN+PCA+\gls{md}} - It follows (1) except that the \gls{md} 
is derived on PCA-reduced features. 

\textit{Results:} Dimensionality reduction (2) drastically improves the network's performance, as shown in the first line of \gls{tab}~\ref{tab:ablation}. The improvement amounts to 7-30\% on the \gls{ood} metrics and 3\% on the \gls{id} metrics.
One possible explanation is that 
the reduced dimensions are the ones that contribute to distinguishing \gls{id} samples from \gls{ood} ones, as previously observed by \cite{ren2021simple}.
When including all the feature dimensions in the \gls{md}, 
the dimensions that do not contribute to discriminating \gls{id} and \gls{ood} samples add up and dominate the final \gls{md} score.

\textbf{Distance Definition.} 
We compare Mahalanobis distance and Euclidean distance (ED) in the network's latent space. 
We compare \textbf{(2) DNN+PCA+MD} with the new experiment \textbf{(3) DNN+PCA+ED} - It follows (2) except that the \gls{md} is replaced with \gls{ed}. As for \gls{md}, the $\chi^2_{d'}$ distribution is used to obtain the probability values from \gls{ed} (Sec~\ref{sec:uncertainty_estimation}).

\textit{Results: }The results show that \gls{md} boosts the performance in terms of \gls{id} and \gls{ood} metrics. The improvement is \gls{ece} score is by 5\%, and the \gls{ood} metrics improved by 3-9\% when using \gls{md}. This is because \gls{md} takes into account the data correlation, which gives a better estimate of the probability and uncertainty values.

\textbf{Representation training.} 
To study the influence of the training on the representations, we consider three experiments: \textbf{(4) DNN+Triplet+PCA+MD} - We train the \gls{dnn} using both cross-entropy and triplet loss. 
\textbf{(5) DNN+Clustering+PCA+MD} - We train using the cross-entropy loss only and periodically cluster the feature points using X-Means.
\textbf{(6) MAPLE} - This is our proposed method that fuses (4) and (5).
For all experiments, the \gls{md} is derived on the reduced features.

\textit{Results: } Using the triplet loss (4)
improves
the performance considerably compared to training with the cross-entropy loss only (2). An explanation is that the triplet loss pulls in-class feature embeddings together, and pushes the other class features apart.
This encourages the representations to be well separated and makes it easier to distinguish \gls{ood} features.
Choosing the triplet loss for metric learning is empirically motivated: experiments using contrastive loss showed that triplet loss has a slightly better performance. 

Periodic clustering (5) improves the \gls{ece} score
by 2\%, and the \gls{auroc} and \gls{aupr} scores on SVHN
by about 7\% compared to (2). However, there is a slight drop in accuracy by 0.3\% and \gls{ood} metric by 4\% on CIFAR100. One explanation is that clustering increases the chances
of new classes to overlap. 
This phenomenon is illustrated
in the centre plot of Fig.~\ref{fig:cluster_example}.
The class overlap is particularly hindering
when the new domain is close to the training one: 
with clustering (5), the SHVN scores are better but the near-\gls{ood} CIFAR100 performs better without clustering (2).

\method uses clustering together with triplet loss and achieves top-performance. The triplet loss reduces the 
overlap introduced with the clustering by pulling apart the newly created classes. With MAPLE, the latent representations are approximately Gaussian and well-clustered resulting in better \gls{md} estimates and superior performance in both \gls{id} and \gls{ood} metrics. Compared to experiment (2), the calibration error drops by 4\%and the \gls{ood} scores improved by 4-11\%.

\textbf{False Negative Ratio $t$.}
We evaluate the influence of the clustering trigger \ie the False Negatives Ratio. We train \method with a range of $t$ values on CIFAR10 (\gls{tab}~\ref{tab:ablation_fnr_cifar10}).

\textit{Results: }
A low value of $t$ results in overclustering, where multiple clusters contain similar images. This further increases the chances of misclassifications, leading to decrease in the metric values. On the other hand, high $t$ values result in underclustering. Note that for $t>0.3$, there are no additional clusters generated. This is because, the classes have false negative ratios that are below this threshold and so, they are not clustered. For CIFAR10, a $t$ value of 0.3 yields the best results. 

An extended ablation analysis on the influence of classification backbones, clustering methods, and hyperparameters is provided in Appendix.~\ref{app:ablation}.

\begin{table}[t]

    \begin{adjustbox}{width=\columnwidth, center}
    \centering
    \begin{tabular}{c c c c c c }\hline
        {False Negative}&{}&{}&{}&{SVHN}&{CIFAR100}\\
        { Ratio (t)}&{\#Classes}&{Accuracy$\uparrow$}&{ECE$\downarrow$}&{AUROC$\uparrow$}&{AUROC$\uparrow$}\\\hline
        0.0&23&0.9449&0.014&0.922&0.888 \\
        0.1&18&0.9534&0.013&0.964&0.918\\
        0.2&14&\textbf{0.9544}&0.012&0.991&0.925\\
        0.3&12&0.9541&\textbf{0.012}&\textbf{0.996}&\textbf{0.926}\\ 
        0.4&10&0.9535&0.013&0.961&0.921\\
        0.5&10&0.9535&0.012&0.955&0.915\\
        \hline
        
    \end{tabular}
    \end{adjustbox}
    \medskip
    \caption{\textbf{Metrics for different values of False Negative Ratio evaluated on CIFAR10} \#Classes refers to the total number of output classes obtained after clustering. A low value of $t$ results in overclustering, whereas a high $t$ fails to detect classes with high variance.}
    \label{tab:ablation_fnr_cifar10}
\end{table}

\section{Discussion} 
With the periodical clustering 
and the dynamic re-labeling, a natural question that arises is \textit{'Is there a drop in performance when the ground truth labels change during training?'}. Experimentally, we observe a drop in training accuracy by 2-3\% in the following epoch after every clustering phase.
However, the network makes up for the drop within 4-5 epochs of training.  

It can happen that the clusters contain very few samples, which introduces label imbalance when classifying.
This is exacerbated when the samples are over-clustered. To mitigate this, we restrict X-Means to only cluster the classes that get misclassified. 
These are the classes with a false negative ratio higher than the threshold $t$. 
Automatic clustering regularization~\cite{caron2018deep,johnson2019billion,larsson2019fine} is left for future work. 

\section{Use Case: Microorganism Classification}

\begin{figure}[h]
    \centering
    \includegraphics[height = 2cm,width=1cm,angle=90]{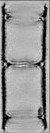}
    \includegraphics[height = 1cm,width=1cm,angle=90]{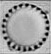}
    \includegraphics[height = 1.5cm,width=1cm,angle=90]{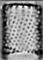}
    \caption{\textbf{Micro-organisms belonging to the same class.} These images of \textbf{one} diatom class show wide appearance changes due to different viewpoints during the acquisition. These translate into separate distributions in the latent space, deviating from Gaussian distribution. MAPLE's regularization makes the latent space Gaussian, hence suitable for MD calculation.  }
    \label{fig:diatoms}
\end{figure}

We consider the real-life computer vision use-case of image-based diatom identification~\cite{du1999diatom}.
Diatoms are microorganisms present in the water. The distribution of diatoms in the water is a useful indicator for predicting the water quality. Diatoms consist of several species or 'taxa', each corresponding to a different class with a different appearance. Typical in several biology applications, the image dataset includes a lot of intra-class variance (Fig.~\ref{fig:diatoms}). In this study, we evaluate the performance of different approaches when encountering taxa that were not previously trained on.

\begin{table}[h]

    \begin{adjustbox}{width=\columnwidth, center}
    \centering
    \begin{tabular}{c c c c c c}\hline
        {Method}&{Accuracy $\uparrow$}&{ECE $\downarrow$}&{AUROC $\uparrow$}&{AUPR $\uparrow$}&{Latency (ms/sample)$\downarrow$}\\\hline
        MC-Dropout~\cite{gal2016dropout}&0.936&0.039&0.548&0.589&129.7\\
        Deep Ensemble~\cite{lakshminarayanan2017simple}&\textbf{0.969}&\textbf{0.025}&0.589&0.570&146.81\\
        SNGP~\cite{liu2020simple}&0.954&0.196&0.798&0.826&26.25\\
        \method&0.963&0.036&\textbf{0.864}&\textbf{0.865}&\textbf{17.38}\\ \hline
        
    \end{tabular}
    \end{adjustbox}
    \medskip
    \caption{\textbf{Real Case Application: microorganism classification.}
        With its top performance and state-of-the-art speed, \method makes for a particularly applicable method for classification and \gls{ood} detection on
        real case datasets.
    }
    \label{tab:diatoms}
\end{table}

We train a Wide ResNet 28-10 on 130 taxa and use 36 taxa as \gls{ood}. The dataset is particularly challenging since it is fine-grained and Near-OOD. Additional details on the dataset and experimental setup are provided in Appendix~\ref{app:diatoms}. As shown in Tab.~\ref{tab:diatoms}, \method outperforms all baselines on \gls{ood} detection. While Deep Ensemble has a slightly better classification accuracy and \gls{ece} score, \method significantly outperforms it in \gls{ood} 
with a 30\% score boost and a runtime 8 times faster.

\section{Conclusion}
This paper presents MAPLE, a self-supervised regularization method for uncertainty estimation and out-of-distribution detection on \gls{cnn} classifiers.
The uncertainty is derived from the Mahalanobis Distance (MD) between an image representation and the class representations in the network's latent space.
\method derives meaningful \gls{md} distances by introducing a regularizer based on self-supervised label refinement and metric learning. Thus, \method learns well-clustered representations that are approximately Gaussian for each class, which complies with the theoretical requirements of \gls{md}-based uncertainty estimation.
Experimental results show that \method achieves state-of-the-art results on out-of-distribution detection 
and is very competitive with existing methods on predictive probability calibration.
\method also has the significant advantage of introducing the least architectural changes.
Finally, we demonstrate a real-life use-case of our method on microorganism classification for the automatic assessment of water quality in natural ecosystems. 

\clearpage
{\small
\bibliographystyle{ieee_fullname}
\bibliography{egbib}
}

\clearpage
\appendix
\section{Metrics Definitions} \label{app:metrics}

In this section, we provide the definitions and formulas of metrics used for evaluation in this paper. Let the samples be represented by $[(x_1,y_1),(x_2,y_2),...,(x_N,y_N)]$, where $N$ is the total number of samples. $x_i$ is the input and $y_i$ is the corresponding label, having values between 1 and $K$.

\textbf{Accuracy.} This gives the fraction of samples that were correctly identified by the network. 

\begin{equation*}
   acc = \frac{1}{N} \sum_{n=1}^N 1[\text{argmax}(p(y_n | x_n))  = y_n]
\end{equation*}
where, $p(y_n|x_n)$ is the predicted probability that the sample $x_n$ belongs to the class $y_n$. A higher accuracy indicates better performance.

\textbf{Expected Calibration Error.} \gls{ece} is a measure of predictive probability calibration error. The output probability is divided into a histogram of $B$ equally spaced bins. The expected calibration error gives the difference between the \textit{observed relative frequency} (accuracy) and the \textit{average predicted frequency} (confidence).
\begin{equation*}
    ECE = \sum_{b=1}^{B}\frac{n_b}{N}|acc(b) - conf(b)|
\end{equation*}
where $n_b$ is the number of samples in bin $b$, $N$ is the total number of samples, $acc(b)$ and $conf(b)$ are the accuracy and confidence of bin $b$. A lower ECE score means that the accuracy and confidence are aligned, indicating better calibration.

\textbf{Negative Log Likelihood.} \gls{nll} calculates the negative log-likelihood for the predicted class probability. While it is generally used for optimization using cross-entropy loss, it is also commonly used to evaluate the prediction uncertainty. A lower NLL score is preferred.
\begin{equation*}
    NLL = \frac{-1}{N}\sum_{n=1}^{N}log(p(y_n|x_n))
\end{equation*}

\textbf{Area Under Receiver Operating Characteristic Curve.} 
\gls{auroc} indicates the ability to separate \gls{id} and \gls{ood} samples. To calculate this metric, the predicted uncertainty is used to determine if a sample is \gls{id} or \gls{ood}. This can be considered as a binary classification problem. The area under the plot between the true positive rate and the false positive rate gives the \gls{auroc} value. Higher \gls{auroc} value means better separation between \gls{id} and \gls{ood}. 

\textbf{Area Under Precision-Recall Curve.} \gls{aupr}, like \gls{auroc} measures the ability to separate \gls{id} and \gls{ood} samples. Considering \gls{id} and \gls{ood} separation as a binary classification problem, the area under the plot between precision and recall values give the \gls{aupr} score.

\section{Experimental details}\label{app:exp}

\subsection{CIFAR10 \vs CIFAR100/SVHN}\label{app:cifar}
CIFAR10~\cite{krizhevsky2009learning} consists of 10 classes. We split the original training set consisting of 50000 samples into train and validation set, in the ratio of 80:20. The validation set was used for hyperparameter tuning. 
The test set consists of 10,000 samples, used for inference. For \gls{ood} analyses, we use the test set of SVHN and CIFAR100, which consists of 26,032 and 10,000 samples respectively. The \gls{ood} images are normalized the same way as train images during inference.

The network architecture is Wide ResNet 28-10~\cite{zagoruyko2016wide}. The feature embedding layer has a dimension of 640. After training \method, the number of classes were 12, and hence, the final layer has a dimension of 12, followed by softmax. We trained the model for 200 epochs. We used an SGD optimizer with a learning rate of 0.05. The momentum was set to 0.9 and weight decay of $1e^{-4}$. The training was performed using PyTorch on a 12Gb NVIDIA GeForce GTX 1080Ti with a batch size of 64. The dimension of the reduced features from PCA is 12. 

\subsection{CIFAR100 \vs CIFAR10/Tiny ImageNet}\label{app:cifar100}
CIFAR100~\cite{krizhevsky2009learning} consists of 100 classes. We split the original training set consisting of 50000 samples into train and validation set, in the ratio of 80:20. The validation set was used for hyperparameter tuning. 
The test set consists of 10,000 samples, used for inference. Additionally, inference and ID metrics were also calculated for the corrupted version (CIFAR100-C~\cite{hendrycks2021natural}). For \gls{ood} analyses, we use the test set of Tiny ImageNet and CIFAR100, which consists of 10,000 samples each. The \gls{ood} images are normalized the same way as train images during inference.

The network architecture is Wide ResNet 28-10~\cite{zagoruyko2016wide}. The feature embedding layer has a dimension of 640. After training \method, the number of classes were 118, and hence, the final layer has a dimension of 118, followed by softmax. We trained the model for 200 epochs. We used an SGD optimizer with a learning rate of 0.05. The momentum was set to 0.9 and weight decay of $1e^{-4}$. The training was performed using PyTorch on a 12Gb NVIDIA GeForce GTX 1080Ti with a batch size of 64. The dimension of the reduced features from PCA is 34. 

\subsection{ImageNet \vs ImageNet-O}
The ImageNet dataset~\cite{imagenet15russakovsky} consists of 1,000 classes with 1,281,167 train, 50,000 validation and 10,000 test images. For \gls{ood} analysis, ImageNet-O~\cite{hendrycks2016baseline} is used, which consists of 200 classes and 2000 images. The \gls{ood} images are normalized the same way as train images during inference.

The ResNet-50~\cite{he2016deep} was used for training. The feature embedding layer has a dimension of 640. After training \method, the number of classes were 1223, and hence, the final layer has a dimension of 1223, followed by softmax. We trained the model for 300 epochs. We used an Adam optimizer with a learning rate of 0.01. The training was performed using PyTorch on a 2 24Gb NVIDIA GeForce RTX 3090 with a batch size of 64. The dimension of the reduced features from PCA is 66.

\subsection{Diatoms}
\label{app:diatoms}
The diatom dataset consists of 9895 individual RGB images of size $256\times256$, belonging to 166 classes~\cite{venkataramanan2023usefulness}. We divide it into \gls{id} dataset consisting of 130 classes (7874 images) and the remaining 36 classes as \gls{ood} (2021 images). 70\% of the \gls{id} images were used for training, 10\% for validation and 20\% for testing. While training, horizontal and vertical flips were used for data augmentation.

The network architecture is Wide ResNet 28-10~\cite{zagoruyko2016wide}. The feature embedding layer has a dimension of 640. After training, there were a total of 158 classes, hence the output layer consists of 158 neuron with a softmax activation. We trained the model for 100 epochs with an Adam optimizer. The learning rate was $2e^{-4}$ and batch size 4. The training was performed using PyTorch on a 12Gb NVIDIA GeForce 1080Ti. The dimension of the features after PCA reduction was 31.

\subsection{Hyperparameter Tuning}\label{app:hyp}
Our training depends on the following  hyperparameters: (1) \textbf{Frequency of epochs $p$} - After every $p$ epochs, validation is performed to obtain the new cluster assignments using X-Means. (2) \textbf{False negative ratio threshold $t$} - $t$ is a threshold used to decide the class features to be clustered. From the normalized confusion matrix obtained during the validation step, the classes having false negative greater than $t$ are clustered using X-Means. (3) \textbf{Maximum number of clusters} - This is a parameter of X-Means, that specifies the upper bound to the number of clusters that X-Means can generate for each class. 

To find the optimal value of these parameters, a grid search was performed. For the grid search, the values of hyperparameters used were: False negative ratio threshold $t \in \{0.0,0.1,0.2,0.3,0.4,0.5,0.6,0.7,0.8,0.9,1.0\}$, frequency of validation epochs $p\in\{5,10,15,20\}$ and maximum number of clusters that X-Means can generate $\{3,5,7,10\}$. 

From the grid-search analysis, the best performance was obtained when $t=0.3$, $p=10$ and maximum number of clusters=5 for CIFAR10, CIFAR100 and the Diatom datasets. For ImageNet, $t=0.2$, $p=20$ and maximum number of clusters=5.

\subsection{Loss Functions}\label{app:loss}
For our training, we use the Cross-Entropy Loss and the Triplet Loss. 

\subsubsection{Cross-Entropy Loss}
To estimate the cross-entropy loss, the final layer of the model is passed through a softmax layer to obtain probability values. Cross-entropy loss increases proportional to the difference between the predicted probability and the actual probability (typically 1) of the ground truth class. The cross-entropy loss is given by:
\begin{equation}
    \mathcal{L}_{\text{cross-entropy}} = -\sum_{i=1}^{K} y_i \log(p_i)
\end{equation}
where $K$ is the total number of samples, $y_i$ is the binary one-hot encoding value corresponding to ground truth class, which equals 1, and $p_i$ is the probability predicted by the network.

\subsubsection{Triplet Loss}
To estimate the triplet loss, we use the feature embedding obtained from the penultimate layer of the classification network. Triplet loss tries to minimize the distance of intra-class data points, while maximizing the inter-class distance. Consider three input samples, which are feature embeddings extracted: anchor $x^\prime_a$, positive $x^\prime_p$ and negative $x^\prime_n$. $x^\prime_a$ and $x^\prime_p$ belong to the same class while $x^\prime_n$ belongs to a different class. The triplet loss is given as:
\begin{equation}
    \mathcal{L}_{\text{triplet}} = \max\{||x^\prime_a-x^\prime_p||-||x^\prime_a-x^\prime_n||+\alpha , 0\}
\end{equation}

The final objective is 
\begin{equation}
    \mathcal{L}_{\text{total}} = \mathcal{L}_{\text{cross-entropy}} + \mathcal{L}_{\text{triplet}}
\end{equation}

\section{Algorithm} \label{app:algorithm}
The proposed method is summarized in Algorithm~\ref{alg:train} and Algorithm~\ref{alg:test}. Algorithm~\ref{alg:train} provides the steps using in training \method. Algorithm~\ref{alg:test} summarizes the procedure for estimating uncertainty from \gls{md}. At regular intervals of the training process, validation is performed, and the train feature representations are clustered using X-Means. The time complexity for X-Means is O(log K), where K is the number of clusters. The train features are reduced in dimension using PCA, which has a complexity of O($nd^2$+$d^3$), where $n$ is the number of train data and $d$ is the feature dimension. Mahalanobis distance calculation requires calculating mean and the covariance matrix, which has a complexity of O($nd'$) and O($d'^3$), where $d'$ is the PCA reduced feature dimension. 

Note that the operations such as the PCA covariance calculation and eigenvalue decomposition, and inverse covariance calculation for \gls{md} is to be performed only once, at the end of the training. During inference, the calculated mean and inverse covariance matrix can be used to calculate the Mahalanobis distance for all the test points.

\begin{algorithm}
\DontPrintSemicolon
\caption{\method training}\label{alg:train}
\SetKwInOut{Input}{Initialize}\SetKwInOut{Output}{Model}
\KwData{
Ground truth labels $\boldsymbol{y} \in \{1,2,...k\}$,\\
Input samples $\boldsymbol{x} \in \mathbb{R}^D$,\\
Train input samples $\boldsymbol{x_{train}}=\{x_n\}_{n=1}^{N}$,\\
Train dataset $\boldsymbol{\mathcal{D}_{train}}=\{(x_n,y_n)\}_{n=1}^N$, \\
Validation dataset $\boldsymbol{\mathcal{D}_{val}}=\{(x_v,y_v)\}_{m=1}^M$

}

\Input{$n_c=k, p=10, t=0.3, max\_clusters=5$}
\Output{$f^\theta:\mathbb{R}^D\rightarrow\mathbb{R}^d$}
\For{epoch = 1 \KwTo \text{max-epochs}}
{
    {Train $f^\theta$ with $\boldsymbol{\mathcal{D}_{train}}$ and $n_c$ classes and loss given by $L_{total}=L_{cross-entropy} + L_{triplet}$} \\
    \If{epoch\%p==0}
    {
        $\boldsymbol{x'_{train}}=f^\theta(\boldsymbol{x_{train}})$\\
        Get softmax predictions on $\boldsymbol{\mathcal{D}_{val}}$ \\ if $n_c>k$, remap pseudo-labels to original class labels \\
        Compute confusion matrix\\

        \For{i=1 \KwTo k}
            {
            \If{false\_negative\_ratio(i) $>$ t}
                {
                Cluster using X-Means.
                X-Means($\boldsymbol{x'_{train}(i)}$, max\_clusters)
                }
            }
        $K \gets $ total number of clusters obtained from all the classes\\
        $n_c = K$ \\
        Update $\boldsymbol{\mathcal{D}_{train}}$ with pseudo-labels from clustering

    }

}

\end{algorithm}

\begin{algorithm}
\DontPrintSemicolon
\SetAlgoLined
\caption{\method Prediction}\label{alg:test}
\SetKwInOut{Input}{Input}
\KwData{
Train feature embeddings $\boldsymbol{x'_{train}}$
}
\Input{Test sample $\boldsymbol{\tilde x}$}
Compute the reduced dimensional train features: $\boldsymbol{z_{train}} = g(\boldsymbol{x'_{train}})$

Compute individual class means and shared covariance $\mu_c, \boldsymbol{\Sigma}$ \\
  $\mu_c = \frac{1}{N_c}\sum_{i:y_i=c} z_i$\\
  $\boldsymbol{\Sigma} = \frac{1}{N}\sum_c\sum_{i:y_i=K}(z_i-\mu_c)(z_i-\mu_c)^T$

Get reduced dimensional feature for $\boldsymbol{\tilde x}$:
$\boldsymbol{\tilde z} = g(f^\theta((\boldsymbol{\tilde x}))$

Compute Mahalanobis distance:
    $MD(\boldsymbol{\tilde x})= \sqrt{(\boldsymbol{\tilde{z}}-\mu_c)^T\boldsymbol{\Sigma}^{-1}(\boldsymbol{\tilde{z}}-\mu_c)}$
    
Get the prediction probabilities:
$P_{MD} = 1 - \text{cdf}(\chi^2_{d^\prime})(MD^2)$\\
Predicted class = argmax($P_{MD}$)

Compute uncertainty
$u = \text{cdf}(\chi^2_{d^\prime})(MD^2)$

\end{algorithm}

\section{Additional Experiments} \label{app:additional_exp}
In this section, we provide results for additional evaluation of \method.

\begin{table}[h]

    \begin{adjustbox}{width=0.75\columnwidth, center}
    \centering
    \begin{tabular}{c c c c}\hline
        {Method}&{acc@.50}&{acc@.80}&{acc@.90}\\\hline
        MC Dropout~\cite{gal2016dropout}& 0.962& 0.976&0.988 \\
        Deep ensemble~\cite{lakshminarayanan2017simple}&\textbf{0.967}&0.987&\textbf{0.995}\\ 
        DUQ~\cite{van2020uncertainty} & 0.950 & 0.977 & 0.982 \\
        SNGP~\cite{liu2020simple} & 0.959 & 0.978 & 0.985 \\
        DUE~\cite{van2021feature} & 0.962 & 0.974 & 0.979 \\
        \method&0.958&\textbf{0.989}&\textbf{0.995}\\ \hline
    \end{tabular}
    \end{adjustbox}
    \medskip
    \caption{\textbf{Accuracy on CIFAR10 with different confidence levels.} \method achieves top accuracy at confidence levels of 0.80 and 0.90. }
    \label{tab:cifar_acc}
\end{table}

\subsection{Accuracy based on prediction confidence} \label{app:accuracy}

We evaluate the accuracy of prediction when selecting samples with predictive confidence above a given threshold. In other words, classification is performed only when the network's confidence is above a threshold. This is representative of real-life applications where a network's prediction is considered only when the confidence is high. 
We consider three probability thresholds: 0.50, 0.80 and 0.90. For all samples with predictive probability above these values, we report the classification accuracy. Table~\ref{tab:cifar_acc} gives the results on the test set of CIFAR10~\cite{krizhevsky2009learning} dataset. 

\textbf{Results.} \method achieves the best accuracy at confidence values of 0.80 and 0.90 on CIFAR10. Overall, on CIFAR10, \method has competitive accuracy with the other approaches. This shows that even though \method is computationally efficient, it can achieve the same level or better performance as the other methods.

\subsection{Gaussian test}
In Section~\ref{sec:train}, it was theoretically shown that X-Means creates clusters of feature points that are Gaussian. In this section, we empirically test this.
A commonly adopted method to check for multivariate Gaussian is to use a quantile-quantile plot, where an observed quantile is compared with a theoretical one. If the samples are Gaussian, their squared \gls{md} follows a $\chi^2$ distribution. Thus, we use $\gls{md}^2_{c^*}$ of the samples feature embeddings as our observed quantile and compare with theoretical $\chi^2$ quantiles.

For our test, we use the reduced feature embeddings, $\boldsymbol{z_{train}}$, from a standard classifier network and \method. The $\gls{md}^2_{c^*}$ of samples are calculated and plotted with $\chi^2$ quantiles with $d'$ degrees of freedom, where $d'$ is the dimension of feature embeddings. We measure the error, which is the mean absolute difference between the two quantiles, to test which method generates feature embeddings that are closer to a Gaussian. In the ideal situation, this value should be zero. The larger the error, the greater is the deviation from a Gaussian distribution.

Table~\ref{tab:gaussian_test} shows the errors computed on feature embeddings from CIFAR10 and CIFAR100 dataset. From the results, MAPLE's error is reduced by over 50\%, which shows that the feature representations of \method are more Gaussian than when using a standard DNN classifier. 

\begin{table}[h]

    \begin{adjustbox}{width=0.75\columnwidth, center}
    \centering
    \begin{tabular}{c c c}\hline
        {Method}&{CIFAR10}&{CIFAR100}\\\hline
        Standard CNN&3.540&4.479\\
        \method&\textbf{1.395}&\textbf{1.982}\\ \hline
        
    \end{tabular}
    \end{adjustbox}
    \medskip
    \caption{\textbf{Mean absolute error between squared \gls{md} and $\chi^2$ distribution.} The lower the error, the more Gaussian are the samples. MAPLE's training generates sample distributions that are approximately Gaussian, fitting with the theoretical framework for \gls{md} calculation. }
    \label{tab:gaussian_test}
\end{table}

\section{Extended Ablation Analyses} \label{app:ablation}
\subsection{\method evaluated on different backbones}
\method is tested on three networks: Wide ResNet 28-10~\cite{zagoruyko2016wide}, ResNet-18~\cite{he2016deep} and EfficientNet-B0~\cite{tan2019efficientnet}. Table~\ref{tab:ablation_arch} gives the quantitative metrics for evaluation on CIFAR10 \vs SVHN and CIFAR100.
While it is expected that the accuracy depends on the architecture used, the calibration and \gls{ood} detection are also influenced by the architecture. Wide ResNet, which has more number of parameters than the other two architectures, learns better feature representations for discriminating each class. As the model parameters decrease, there are overlapping feature points between different classes, which explains the lower accuracy and worse calibration and \gls{ood} metrics.
\begin{table}[h]

    \begin{adjustbox}{width=\columnwidth, center}
    \centering
    \begin{tabular}{c||c c|c c }\hline
        {}&{}&{}&{SVHN}&{CIFAR100}\\
        {Architecture}&{Accuracy $\uparrow$}&{ECE $\downarrow$}&{AUROC $\uparrow$}&{AUROC $\uparrow$}\\\hline
        Wide ResNet 28-10~\cite{zagoruyko2016wide}&\textbf{0.954}&\textbf{0.012}&\textbf{0.996}&\textbf{0.926}\\
        ResNet-18~\cite{he2016deep}&0.945&0.029&0.979&0.886\\
        EfficientNet-B0~\cite{tan2019efficientnet}&0.902&0.035&0.942&0.893\\ \hline
        
    \end{tabular}
    \end{adjustbox}
    \medskip
    \caption{\textbf{MAPLE evaluated on different architectures.} The metrics improve as the model parameters increase, suggesting that the network learns better discriminative feature representations, thereby improving the performance.}
    \label{tab:ablation_arch}
\end{table}

\subsection{Evaluation of different clustering methods}
We analyse the performance of \method on CIFAR10 when clustering is performed using K-Means, G-Means~\cite{zhao2009g} and X-Means~\cite{pelleg2000x}. The value of K in K-Means is set to 3. \gls{tab}~\ref{tab:ablation_cluster} shows the results obtained. Based on the results, X-Means yields the best performance. K-Means and G-Means causes overclustering, which leads to worser performance on \gls{ood} detection. Using X-Means, we choose the optimal number of clusters, which performs superior to the others.

\begin{table}[h]

    \begin{adjustbox}{width=\columnwidth, center}
    \centering
    \begin{tabular}{c||c c c|c c }\hline
        {}&{}&{}&{}&{SVHN}&{CIFAR100}\\
        {Clustering method}&{\#Classes}&{Accuracy$\uparrow$}&{ECE$\downarrow$}&{AUROC$\uparrow$}&{AUROC$\uparrow$}\\\hline
        K-Means & 30 & 0.952 & 0.154 & 0.871 & 0.850\\
        G-Means & 67 & 0.910 & 0.266 & 0.710 & 0.627\\
        X-Means & 12 & \textbf{0.954} & \textbf{0.012} & \textbf{0.996} & \textbf{0.926}\\
        \hline
        
    \end{tabular}
    \end{adjustbox}
    \medskip
    \caption{\textbf{Metrics for different frequency of validation epoch} \#Classes refers to the total number of output classes obtained after clustering. K-Means and G-Means lead to overclustering, whereas using X-Means, the optimal number of clusters are generated leading to better performance.  }
    \label{tab:ablation_cluster}
\end{table}

\subsection{Effect of maximum number of clusters}
\gls{tab}~\ref{tab:ablation_max_clusters} shows the results when the maximum number of clusters that can be generated for every class by X-Means is varied, along with different values of false negative ratio $t$ for CIFAR10. For $t>0.5$, none of the classes are clustered, and hence we do not include them. From the results, when the maximum number of clusters are low, \method fails to capture all the within-class variances, whereas higher values result in overclustering. With the maximum number of clusters as 5, \method achieves the best performance.

\begin{table}[]
    \begin{adjustbox}{width=\columnwidth, center}
    \centering
    \begin{tabular}{c|c|ccccc} \hline
    {Max. number} & {} & {}&{} &{}&{SVHN} &{CIFAR100} \\
    {of clusters} & {t} & {\#Classes} & {Accuracy$\uparrow$} & {ECE$\downarrow$}&{AUROC$\uparrow$} &{AUROC$\uparrow$}\\ \hline
    & {0.1} & {14} & {0.9542} & 0.012 & 0.996 & {0.925}\\
    3 & {0.3} & {10} & {0.9540} & 0.014 & {0.972} & {0.919}\\
    & {0.5} & {10} & {0.9533} & 0.012 & {0.958} & {0.917}\\ \hline
    & {0.1} & {18} & {0.9534} & 0.013 & {0.964} & {0.918}\\
    5 & {0.3} & {12} & {0.9541} & 0.012 & {\textbf{0.996}} & {\textbf{0.926}}\\
    & {0.5} & {10} & {0.9535} & 0.012 & {0.955} & {0.915}\\ \hline
    & {0.1} & {18} & {0.9537} & 0.013& {0.959} & {0.894}\\
    7 & {0.3} & {13} & {\textbf{0.9545}} & 0.012 & {0.992} & {0.921}\\
    & {0.5} & {10} & {0.9531} & 0.013 & {0.944} & {0.911}\\ \hline
    & {0.1} & {26} & {0.9519} & 0.014 & {0.909} & {0.863}\\
    10 & {0.3} & {22} & 0.9521 & 0.013 & {0.918} & {0.886}\\
    & {0.5} & {11} & {0.9534} & 0.012 & {0.952} & {0.908}\\ \hline
    
    \end{tabular}
    \end{adjustbox}
    \medskip
    \caption{\textbf{Effect of maximum number of clusters per class on MAPLES's performance.} A high value of cluster numbers causes overclustering whereas a low value does not generate enough clusters. A value of 5 results in optimal number of clusters for \method to learn meaningful representations. }
    \label{tab:ablation_max_clusters}
\end{table}

\subsection{Effect of frequency of validation epochs.}
\gls{tab}~\ref{tab:ablation_epochs} summarizes the metrics for CIFAR10 when the number of epochs after which the validation and cluster refinements are performed is varied. A low value of validation epochs does not give the network enough time to learn representations for the new clusters generated. Whereas, with larger number of epochs, the number of cluster refinements are low. In both these situations, the network does not identify the optimal clusters. \method gives the best results when the validation is performed every 10 epochs.

\begin{table}[t]

    \begin{adjustbox}{width=\columnwidth, center}
    \centering
    \begin{tabular}{c||c c c|c c }\hline
        {}&{}&{}&{}&{SVHN}&{CIFAR100}\\
        {Validation epochs}&{\#Classes}&{Accuracy$\uparrow$}&{ECE$\downarrow$}&{AUROC$\uparrow$}&{AUROC$\uparrow$}\\\hline
        5 & 16 & 0.895 & 0.025& 0.914 & 0.876\\
        10 & 12 & 0.954 & 0.012 & \textbf{0.996} & \textbf{0.926}\\
        15 & 12 & \textbf{0.955} &0.012 & 0.987 & 0.922\\
        20 & 10 & 0.953 &0.013& 0.968 & 0.917\\
        \hline
        
    \end{tabular}
    \end{adjustbox}
    \medskip
    \caption{\textbf{Metrics for different frequency of validation epoch} \#Classes refers to the total number of output classes obtained after clustering. With lower validation epochs, the clustering is too frequent for the network to learn meaningful representations. At lower frequency, the number of cluster refinements are not sufficient.}
    \label{tab:ablation_epochs}
\end{table}

\end{document}